\documentclass[a4paper,11pt]{article}
\usepackage{ifpdf}
\usepackage[top=2.54cm,bottom=2.54cm, left=2.54cm, right=2.54cm]{geometry}
\usepackage{amsmath}
\usepackage{amsfonts}
\usepackage{amssymb}
\usepackage{graphicx}
\usepackage{bm}

\usepackage{enumerate}
\usepackage{color}
\usepackage{float}
\usepackage[font=small,labelfont=bf]{caption} % Required for specifying captions to
\usepackage{booktabs} % Top and bottom rules for table
\usepackage{diagbox}
\usepackage{lscape}
\usepackage{algorithm}
\usepackage{algorithmic}
\usepackage{float}
\usepackage{hyperref}
\hypersetup{colorlinks,allcolors=black}

\title{Machine Learning Approach and Extreme Value Theory to Correlated Stochastic Time Series with Application to Tree Ring Data}
\author{ Omar Alzeley and Sadiah Aljeddani\\
Department of Mathematics, Umm Al-Qura University, Al-Qunfudah University College\\
\href{mailto:}{oazeley@uqu.edu.sa}\\
Department of Mathematics, Umm Al-Qura University, Al-Lith University College \\ 
\href{mailto:}{smgadany@uqu.edu.sa} \\
}
\date{January  2023}

\begin{document}

\maketitle
\begin{center}
{\bf Abstract}
\end{center}
The main goal of machine learning (ML) is to study and improve mathematical models which can be trained with data provided by the environment to infer the future and to make decisions without necessarily having complete knowledge of all influencing elements. In this work, we describe how ML can be a powerful tool in studying climate modeling. Tree ring growth was used as an implementation in different aspects, for example,  studying the history of buildings and environment. By growing and  via the time, a new layer of wood to beneath its bark by the tree. After years of growing, time series can be applied via a sequence of tree ring widths. The purpose of this paper is to use ML algorithms and Extreme Value Theory in order to analyse a set of tree ring widths data from nine trees growing in Nottinghamshire. Initially, we start by exploring the data through a variety of descriptive statistical approaches. Transforming data is important at this stage to find out any problem in modelling algorithm. We then use algorithm tuning and ensemble methods to improve the k-nearest neighbors (KNN) algorithm. A comparison between the developed method in this study ad other methods are applied. Also,  extreme value of the dataset will be more investigated. The results of the analysis study show that the ML algorithms in the Random Forest method would give accurate results in the analysis of tree ring widths data from nine trees growing in Nottinghamshire with the lowest Root Mean Square Error value. Also, we notice that as the assumed ARMA model parameters increased, the probability of selecting the true model also increased. In terms of the Extreme Value Theory, the Weibull distribution would be a good choice to model tree ring data.

\section{Introduction and Related Work}
In the literature, \cite{Douglass1940} developed the statistical methods of tree ring dating according to the analysis of tree ring growth patterns. Often, trees show annual growth increments, where the factors that effect their formation could be represented by analytical parameters ensuring that one ring can differ from others \cite{Sch1988}. Time series can be a suitable procedure for studying the change in such a phenomenon over time. These are well specified if they are stable and have any significant seasonality that needs to be modelled. The utilities of dendrochronology is very common in the environmental sciences; specially, in climate change topics \cite{Stahle1998} cited in \cite{Liudmila2008}. There are various statistical techniques used in calibration methods. Calibration methods can be classified according to a hierarchical structure of complexity; for example, the simple regression analysis for one level and one variable of tree growth. Variations in the indices of tree growth parameters at a specific site are attributed to specifics specific  climatic variables, for example,  total summer precipitation or mean summer temperature \cite{fritts2012tree}. 
Multiple linear regression analysis (MLR) for two levels and $n$ variables of tree growth has been employed, as has principal component analysis (PCA) for two levels and $n^p$ variables of tree growth. Furthermore, orthogonal spatial regression (PCA+ MLR) and canonical regression analysis have also been employed.

In the PCA analysis, correlated data will be transformed into a few independent principle components that represent the original data \cite{J2016}. PCA, apart from reducing the number of potential predictors, also considerably simplifies multiple regression \cite{bradley1995climate, ch2019}. In the PCA, an orthogonal transformation will remove the high dimensional space to lower one without any correlation. In this paper, we use cross-validation transformation to ensure that the resultant samples will have uncorrelated observations. It is well known that, the first PC had the biggest variance and the PCs  were eigenvectors of the symmetric covariance matrix \cite{J2016}.

The earliest spatial dendroclimatic reconstruction used canonical regression analysis to establish the relationship between patterns of tree ring growth and other climate variable patterns. Spatial regression requires the most complex procedures used in climatic reconstruction. Standardized ring-width measurements from trees on spatial arrays of sites can provide continuous yearly paleoclimatic records, and this information can be transferred to estimates of climatic variables and mapped to reveal spatial variations in climate \cite{bradley1995climate}. Canonical regression analysis was the earliest technique used in spatial climatic reconstruction to establish the link between patterns of tree-ring growth and mean sea level patterns \cite{fritts1971multivariate}.

The theory of Extreme value (EVT) was utilized in different fields such as hydrology, meteorology, finance, insurance, and environmental studies \cite{penalva2013topics}. The theory for independent random variables has been fully developed and, today, is well presented in many textbooks. The asymptotic theory of extreme values sample distribution has been developed under the arguments that take the Central Limit Theory into account.. In 1927, Frechet obtained the asymptotic distribution of the maximum. The first study on the extreme limit problem was presented by Fisher and Tippt (1982) and von Mises (1936). The statistical investigation of extreme values has experienced a remarkable growth since the field first began to be investigated about eighty years ago. In this research, three EVT's investigations related to tree ring width problems are presented. Two are novel implementations of extreme techniques to tree ring width studies, while the other is an improvement of extremes theory, as motivated by environmental issues.

Machine learning is a form of artificial intelligence (AI) that helps systems learn on their own by observing and adapting to past experiences. Machine learning is the process of creating computer programs that can learn from data. Learning process starts with observing or recording data as examples, or instruction, in order to find patterns and make better decisions in the future based on what has been observed. The goal is to let computers learn on their own and take appropriate action based on this learning.

Regarding the former studies, a study by \cite{Bhuyan2017} illustrates the differences between ring width index (RWI) and normalized difference vegetation index (NDVI) time series on varying timescales and spatial resolutions, hypothesizing positive associations between RWI and current and previous-year NDVI at 69 forest sites scattered across the Northern Hemisphere.Their results included a model using high-spatial-resolution NDVI time series indicating in a higher proportion of the variance in RWI than that of a model using coarse-spatial-resolution NDVI time series. 

Another study \cite{Mak2020} built a computer program to count tree rings precisely. They presented an automated procedure for the full pipeline of tree ring analysis.

Both studies differ from the current study as they did not utilize the machine learning algorithm in either counting the tree rings or in showing the differences between two different tree indexes' time series. The current study uses a machine learning algorithm to determine the true model in which the relation between the effects of climate change and tree ring width in the available datasets can be explained. The implementation in this study shows accurate methodology to analyse tree ring width data from nine trees growing in Nottinghamshire using ML algorithms. The rest of this paper can be summarized as follows. Section 2 introduces and prepares tree ring data for machine learning with preprocessing. Section 3 discusses the simulation of correlated stochastic time series. Section 4 estimates the model accuracy by using the resembling method and finding best algorithm for the problem; the tune machine learning algorithm is also discussed. Section 5 models the data using the Extreme Values Theory. Finally, in Section 6, we present a discussion of a few key points, including hints at possible future work.

\newpage
\section{Tree Ring Data}

The data was collected from Thoresby Estate, Sherwood Forest, in Nottinghamshire in 1975 by members of the Nottingham University Tree-ring Dating Laboratory \cite{laxton1982tree}. The data consists of the ring widths of nine trees from the site. Details of the trees are given in Table \ref{table1}. Four of the trees were felled in 1975 and the remaining five in 1967. The dates of the first rings measured vary from 1810 for THO-A01B to 1835 for THO-B05A.
\\
\begin{table}[H]
\captionof{table}{Tree ring data from Thoresby, Nottinghamshire}
\begin{tabular*}{0.95\linewidth}{@{\extracolsep{\fill}}  c | c | c | c  }
\hline
Sample &Number of rings &Date of first ring   & Date of last ring  \\
\hline
THO-A01B&166 &1810 &1975 \\
\hline
THO-A02A & 157 & 1819  & 1975  \\
\hline
THO-A03A &  154  & 1822 &1975 \\
\hline
THO-B01A &  157 & 1820 & 1976\\
\hline
THO-B02B & 157   & 1820 &1976 \\
\hline
THO-B03B &  155  & 1822 &1976 \\
\hline
THO-B04A &  143  &1834  & 1976\\
\hline
THO-B05A & 142  & 1835&1976 \\
\hline
THO-O01C & 144   & 1832 & 1975\\
\hline
\end{tabular*}
\label{table1}
\end{table}
In order to carry out our statistical investigation, we should consider the years in which every tree has one ring. From Table \ref{table1}, we note  that this corresponds to the years 1835 to 1975 (inclusive). 
 We modified the data to start from 1835 to 1975, for example, THO-A01B in the original series data went from 1810 to 1975 with 166 rings. We assume that the start date is 1835 and the end date is 1975, with 141 rings. 
We applied this rule to all nine trees in the dataset. Therefore, we will have fixed and equal numbers of rings for all trees, which is 141, as can be seen in Table \ref{table2}.
\\
\begin{table}[H]
\captionof{table}{Tree ring data for the years 1835 to 1975}
\begin{tabular*}{0.95\linewidth}{@{\extracolsep{\fill}}  c | c | c | c  }
\hline
Sample & Number of rings & Date of first ring   & Date of last ring  \\
\hline
THO-A01B&141 &1835 &1975 \\
\hline
THO-A02A & 141 & 1835  & 1975  \\
\hline
THO-A03A &  141  & 1835 &1975 \\
\hline
THO-B01A &  141 & 1835 & 1975\\
\hline
THO-B02B & 141   & 1835 &1975 \\
\hline
THO-B03B &  141  & 1835 &1975 \\
\hline
THO-B04A &  141  &1835  & 1975\\
\hline
THO-B05A & 141 & 1835&1975 \\
\hline
THO-O01C & 141   & 1835 & 1975\\
\hline
\end{tabular*}
\label{table2}
\end{table}

\section{Simulation of Correlated Stochastic Time Series}
We generate 200 datasets from the assumed true model using ARMA(1,1). We assume 11 true models from ARMA (1,1). We then apply the ML algorithm to estimate the simulated 11 models. We use the Akaike Information Criterion (AIC) as a model selection criterion with the true assumed parameters $\theta$ and $\phi$, where $\theta$ represents the parameter of the autoregressive model and $\phi$ represents the parameter of the moving average model. We count the number of times that the AIC selects the true model from the simulated data. We record the proportion of the times that the AIC succeeded in selecting the true model, as reported in Table \ref{table3}. 
%A simulation study was performed to see the performance of the Akaike Information Criterion (AIC) as a model selection criterion. 
The steps to the methodology can be described as follows:
\begin{enumerate}
\item We generated 140 observations from ARMA(1,1) with some specific parameter values $\theta$ and $\phi$ for the AR and MA, respectively. 
\item We then fitted 11 ARMA models and calculated the AIC for each of them. 
\item We repeated steps 1-2 200 times. 
\item Finally, we calculated the proportion of times the AIC identified the true model. 
\item We repeated steps 1-4 for specific parameter values $\theta$ and $\phi$ for the AR and MA, respectively. 
\end{enumerate}

\begin{table}[H]
\caption{The percentage of true models selected by the AIC}
\begin{tabular*}{0.95\linewidth}{@{\extracolsep{\fill}} l| |l| |l| |l| |l|  |l }
\hline
\diagbox{$\theta$}{$\phi$} & 0.1 & 0.3 & 0.5& 0.7& 0.9 \\
\hline
0.1 & 0.095 & 0.030 & 0.055 & 0.090 & 0.135 \\
\hline
0.3 &  0.015& 0.095& 0.340 &0.445& 0.485  \\
\hline
0.5 & 0.080 & 0.335 & 0.515 & 0.595 & 0.710 \\
\hline
0.7 &  0.130& 0.470& 0.635& 0.760& 0.855\\
\hline
0.9 &  0.200 & 0.560 & 0.695& 0.870& \boldmath{0.970} \\
\hline
\end{tabular*}
\label{table3}
\end{table}
\begin{figure}[H]
\centering
\includegraphics[scale=0.8]{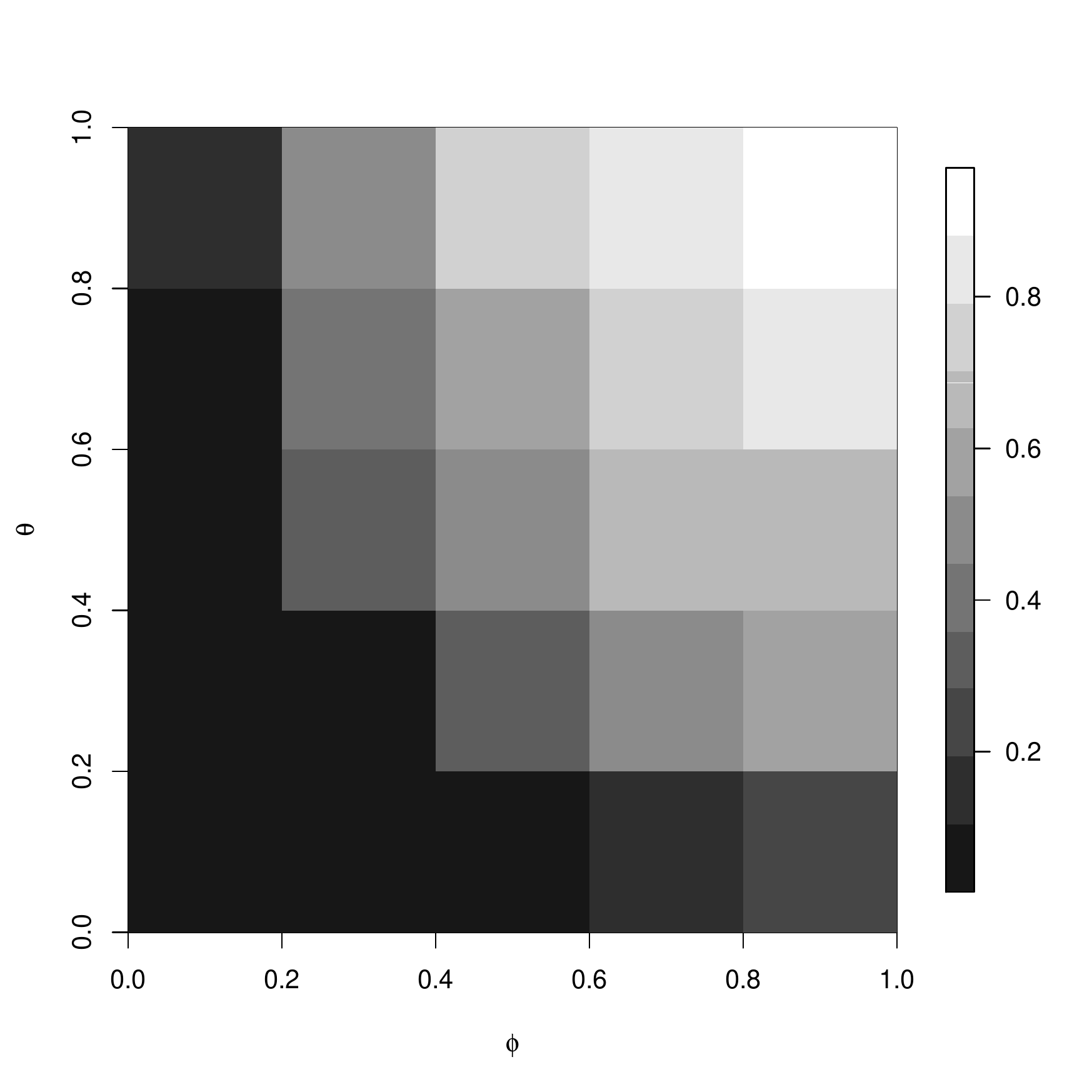}
\caption{Image plot of simulation result}
\label{s1}
\end{figure}
The results of the simulation study are summarised in Table \ref{table3}, 
and also appear in the image plot in Figure \ref{s1}. As shown in Table \ref{table3}, when the parameter values are low, the estimated number of times that the AIC identifies the true model is very small. However, as the parameter values increase, the ability of the AIC to identify the true model increases. For example, the AIC selected the true model $9\%$ of the time with the parameters (0.1, 0.1), while it succeed in selecting the true model (0.9, 0.9) $97\%$ of the time.
\clearpage
\section{Machine Learning Approach to Correlated Stochastic Time Series}
The accuracy of the models estimated via the ML algorithm can be measured by the k-nearest neighbors (KNN) algorithm. Nearest neighbors can be defined as those data points that have the minimum minimum  distances in feature space from the simulated data point, whilst $K$ is the number of such data points that we consider in the implementation of the ML algorithm \cite{S2019}. The six algorithms (linear and nonlinear) have been evaluated using the KNN algorithm. It can be seen from Table \ref{table4} that the KNN value has the lowest Root Mean Square Error of 4.220. We then used a feature selection and removed the most correlated attributes (highly correlated). However, we can see that removing the highly correlated attributes yielded a worse Root Mean Square Error for the linear and nonlinear algorithms. %Thus, the correlated attributes that we removed are contributing to the accuracy of the models.
We notice that some of the trees have a skew and others perhaps have an exponential distribution, so we decided to use the Box-Cox transformation. It can be seen that this decreases the Root Mean Square Error for the linear algorithms only. The accuracy of KNN has been improved by tuning their parameters. We design a grid search around a $K$ value of $6$; this is done by assuming a grid value between $2$ and $10$. Ensemble methods have been used for the problem as this further decreases the value of the Root Mean Square Error. It is noticeable that the random forest was the most accurate algorithm, with a lower Root Mean Square Error than that achieved by tuning the KNN algorithm. 

\begin{table}[H]
\centering
\captionof{table}{The Root Mean Square Error accuracies of the estimated models}
\begin{tabular}{l l l l}
\toprule
\textbf{Algorithms} & \textbf{Row }& \textbf{Transformed } & \textbf{Modified }\\
 & \textbf{dataset }& \textbf{ dataset} & \textbf{dataset }\\

\midrule
Linear Regression &  5.619& 5.374 & 5.636 \\
Generalized Linear Regression &  5.619&5.374  &5.636\\
Penalized Linear Regression & 5.915 &5.544 &5.674 \\
Support Vector Machine &  4.634 & 4.661 & \boldmath{4.185}  \\
Classification and Regression Tree & 5.399 &5.266 &5.266  \\
k-Nearest Neighbor &\\boldmath{4.220}  &5.748 & 5.248 \\
\bottomrule
\end{tabular}
\label{table4}
\end{table}

\begin{center}\vspace{1cm}
\includegraphics[scale=0.6]{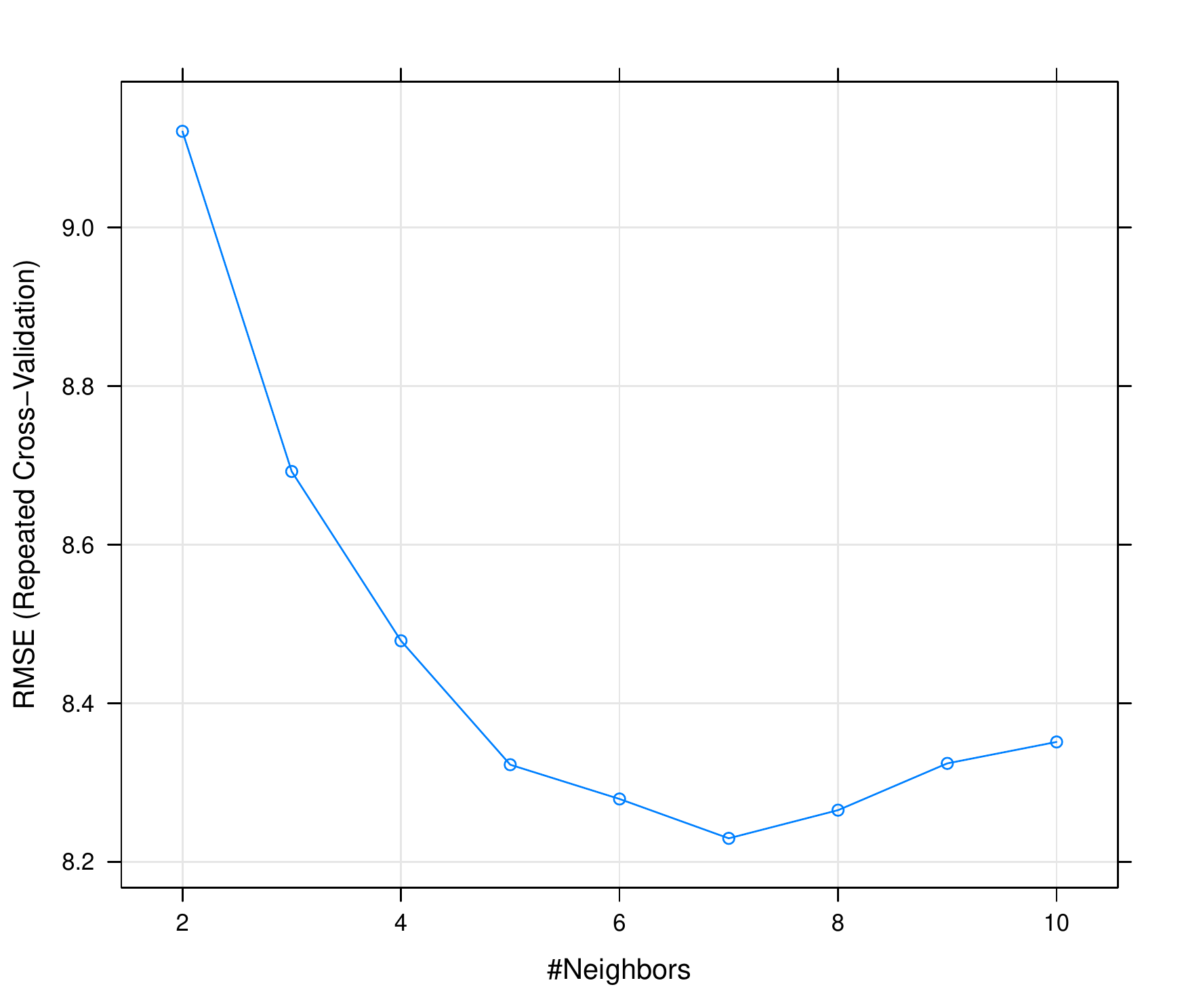}
\captionof{figure}{Algorithm tuning results for KNN on tree ring dataset}
\end{center}\vspace{1cm}

\begin{center}\vspace{1cm}
\includegraphics[scale=0.6]{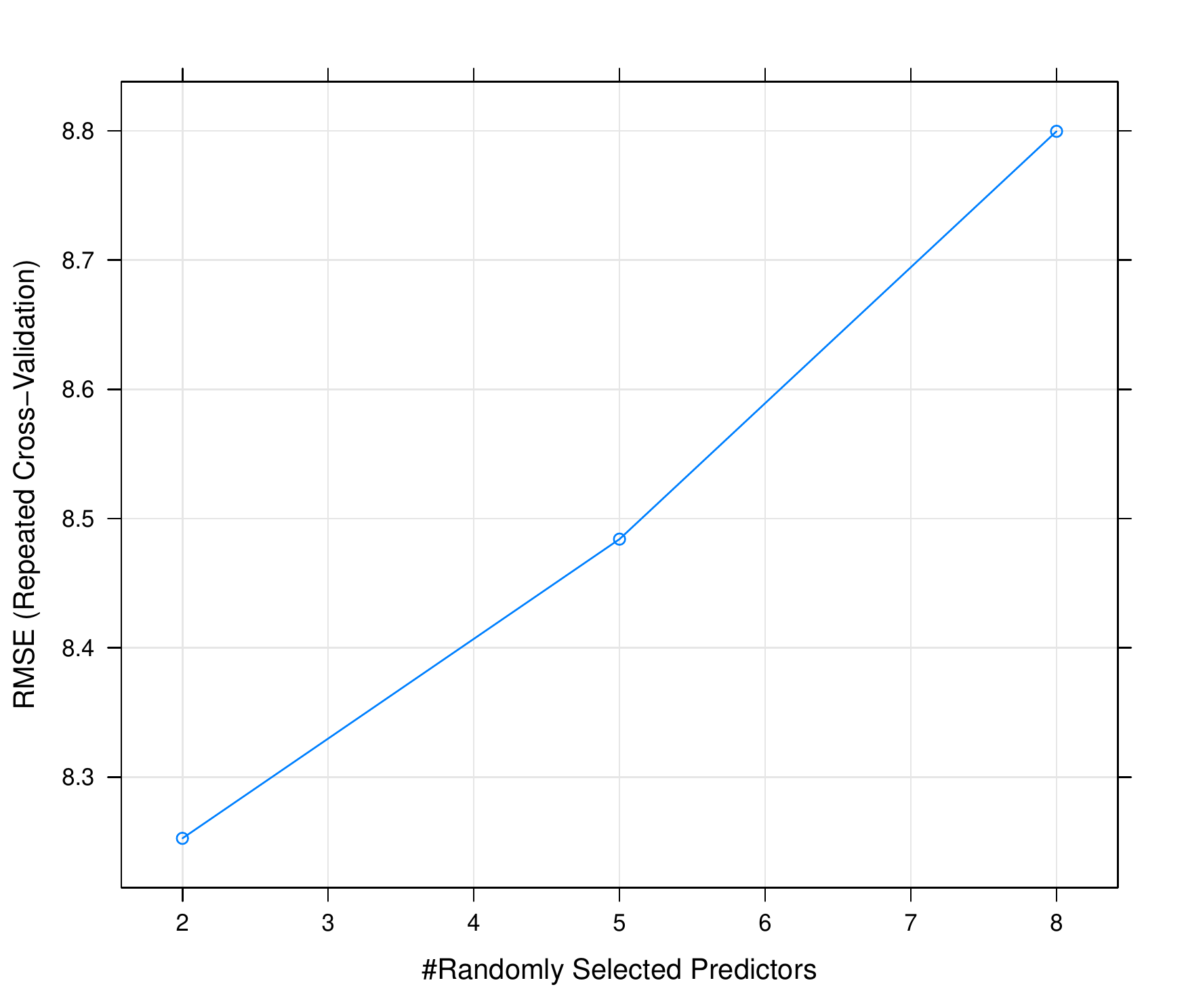}
\captionof{figure}{Tuning the parameters of the random forest on the tree ring dataset}
\end{center}\vspace{1cm}

\newpage
\section{Modeling of Extreme Values in Tree Ring Dating} 

Consider the sample $(X_1, \dots , X_n) \overset{\mathrm{iid}}{\sim}  F$, and assume that we are interested in deriving the distribution of the maximum $M_n$. It is known that 
\begin{equation}\nonumber
P \Big[M_{n} \leq x \Big]= P\Big[X_{1} \leq x,\dots,X_{n} \leq x \Big ]= P[X_{1} \leq x] \dots P \Big[X_{n} \leq x \Big]= F^{n} (x).
\end{equation}
Let $X_1$, $X_2$,$\dots$, $X_n$ be a series of independent random variables with common distribution function $F$
\begin{eqnarray}\nonumber
\lim_{n\to\infty}P\left( \dfrac{M_n - b_n}{a_n}\leq x\right)  =\lim_{n\to\infty} F^{n} \Big (a_{n} x+b_{n} \Big)= G(x) \; \quad \forall \; x \in R
\end{eqnarray}
where $G(X)$ is a nondegereate distribution function. This function is called the Extreme Value and is given by 
\[
EV_{\gamma}(x) = \left\{ \begin{array}{ll}
\exp\Big [-(1+ \gamma x)^\frac{-1}{\gamma} \Big ], \quad 1+ \gamma x >0 & \mbox{if $\gamma\neq0$}\\
\exp \Big [-\exp(-x) \Big ], \quad x \in R , & \mbox{if $\gamma=0$}.\end{array} \right. \]
The external type theorem says that $G(x)$ must be a Gumble, Frechet, or Weibull distribution.

\begin{figure}[H]
\centering
\includegraphics[scale=0.5]{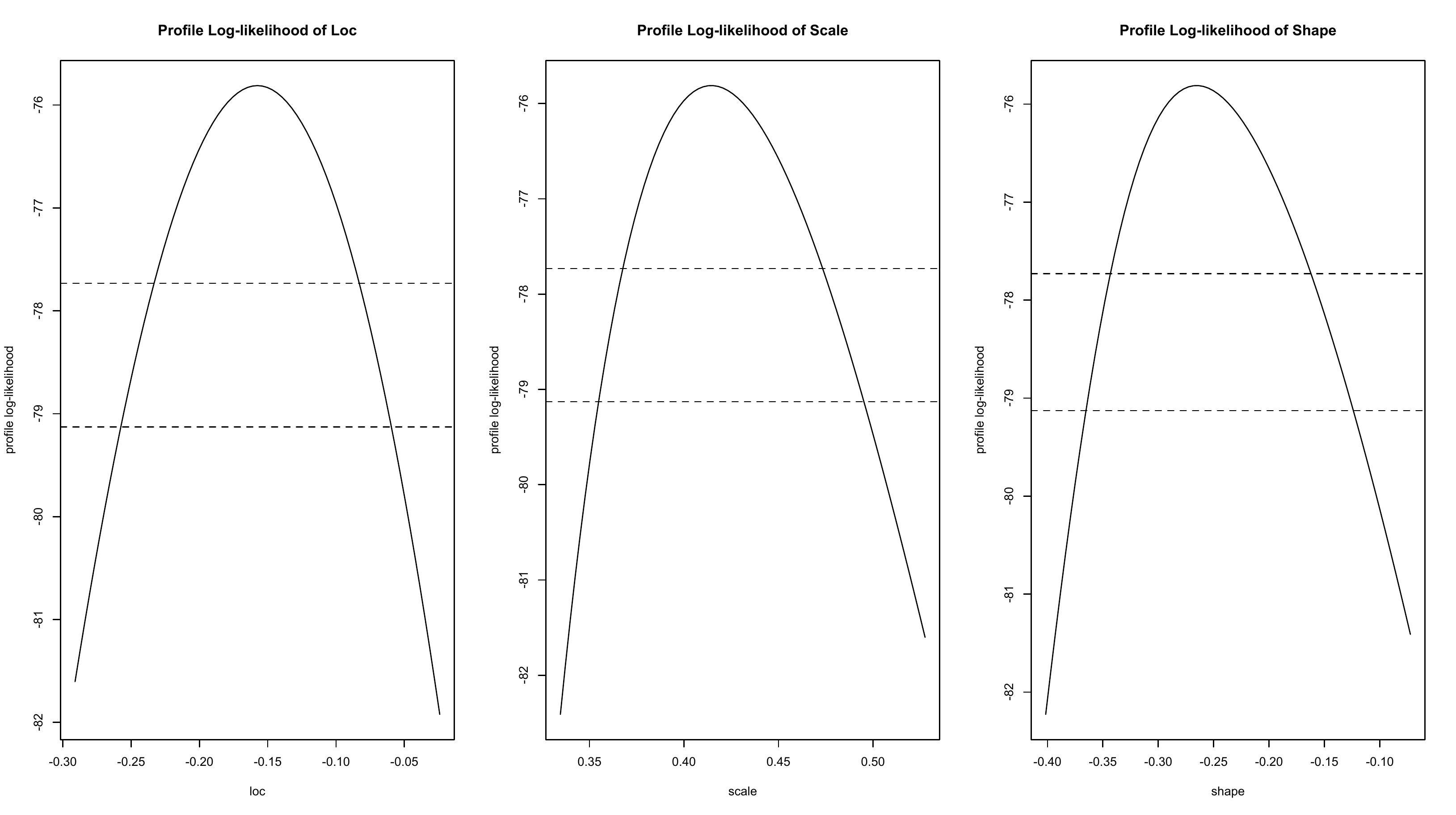}
\caption{Profile log-likelihood for the three parameters in THO-A01B}
\label{Fig.4}
\end{figure}

\begin{figure}[H]
\centering
\includegraphics[scale=0.5]{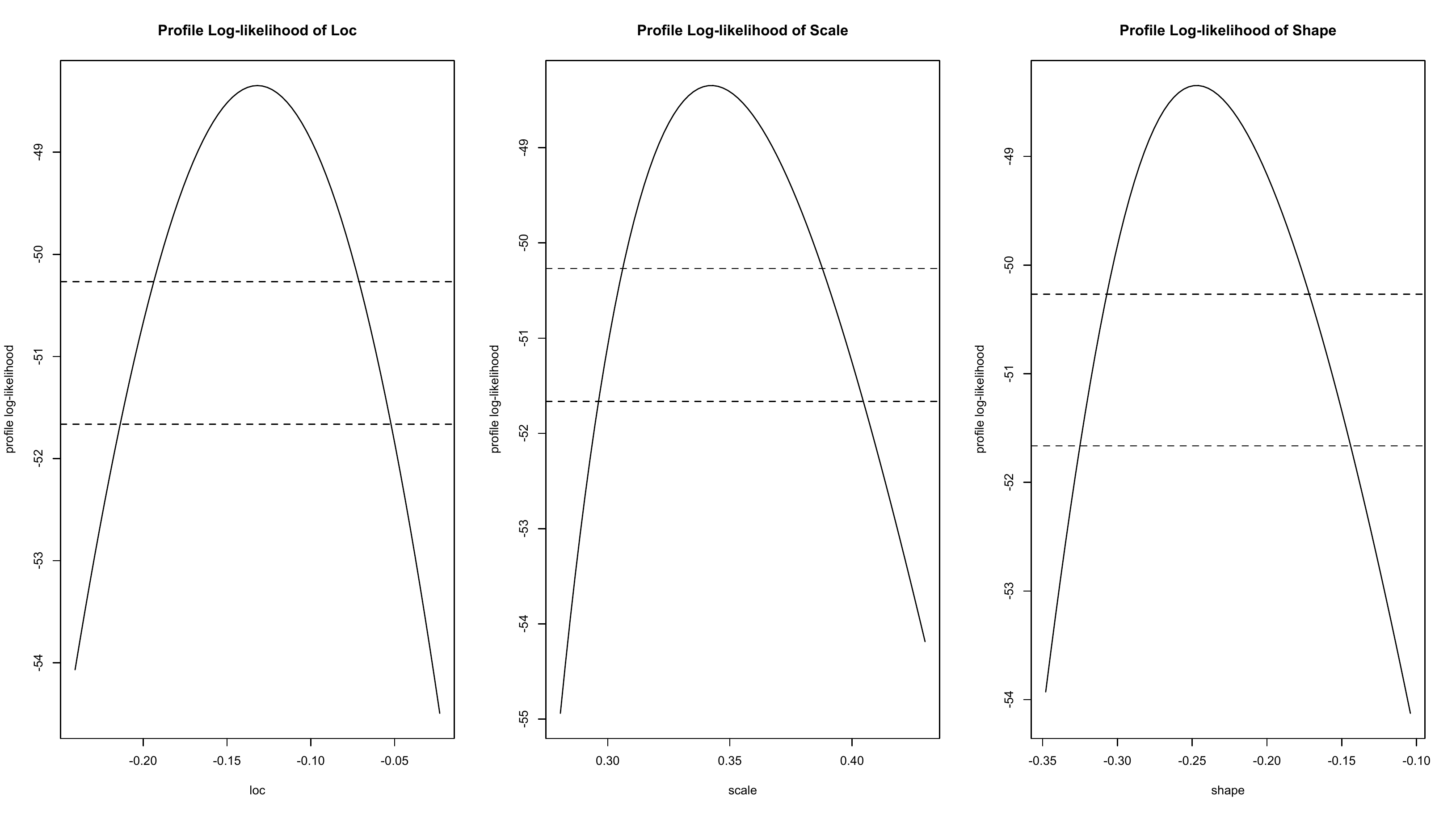}
\caption{Profile log-likelihood for the three parameters in THO-A02A}
\label{Fig.5 }
\end{figure}

\begin{figure}[H]
\centering
\includegraphics[scale=0.5]{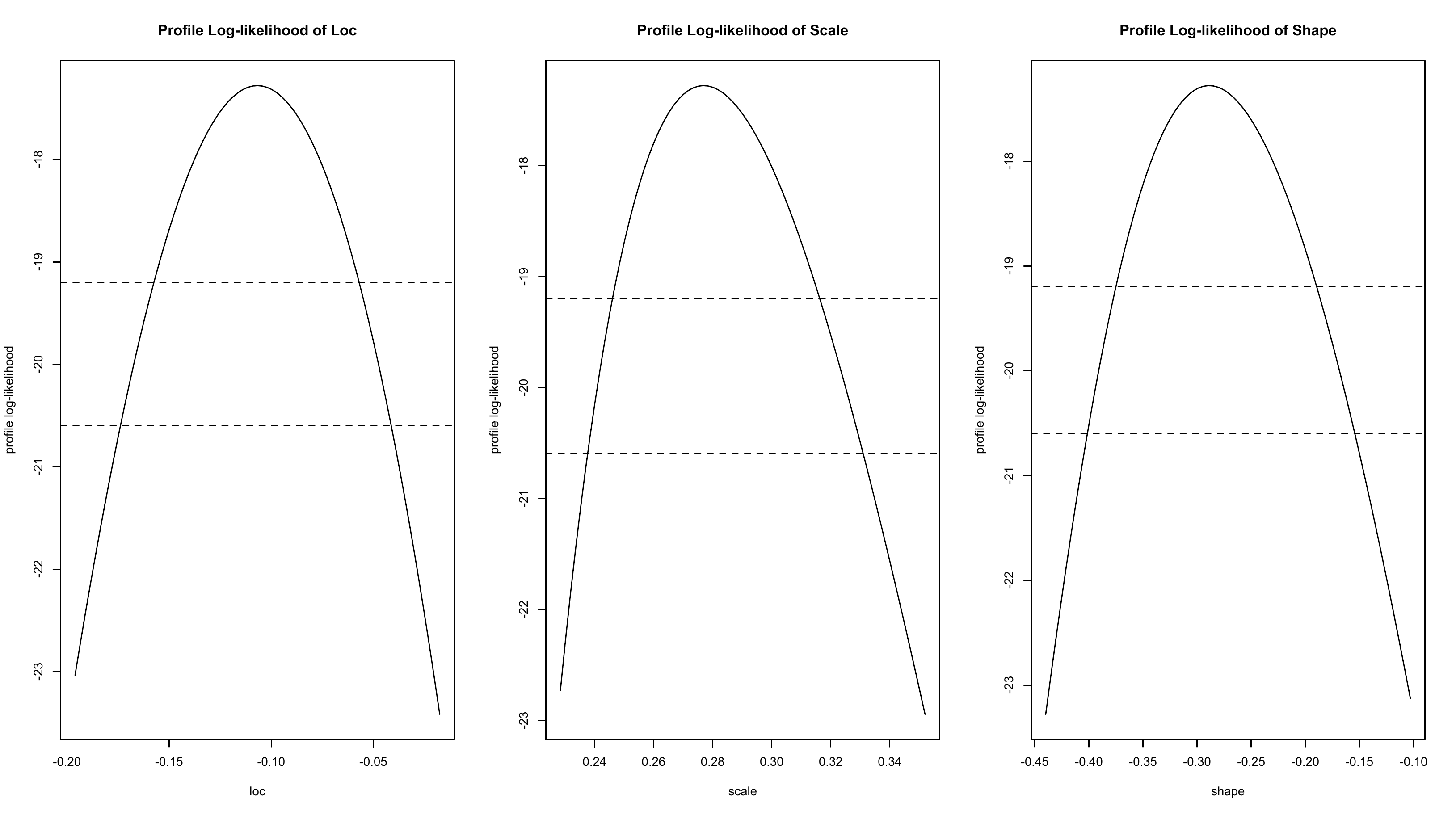}
\caption{Profile log-likelihood for the three parameters in THO-A03A}
\label{Fig.6 }
\end{figure}

\begin{figure}[H]
\centering
\includegraphics[scale=0.5]{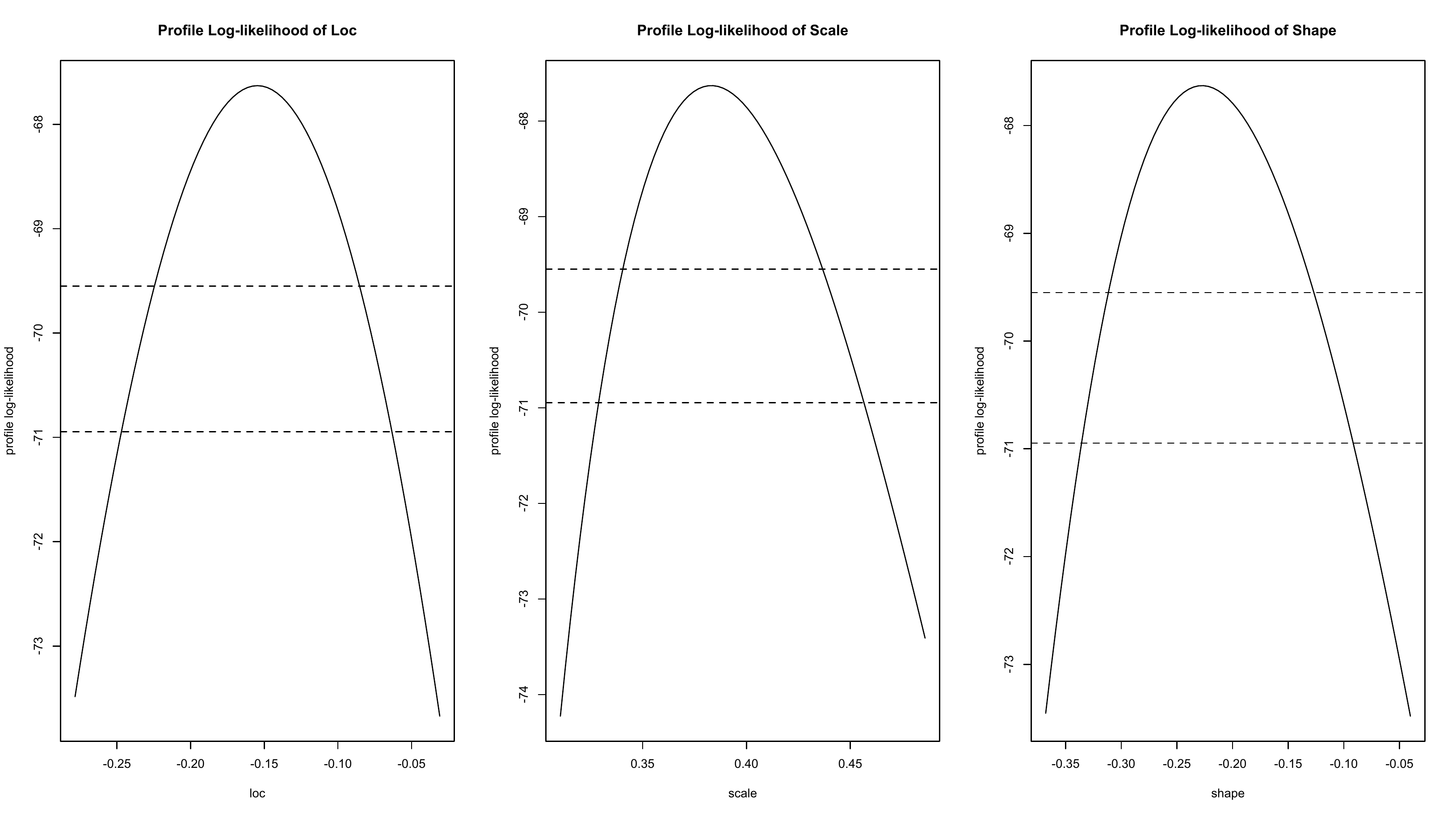}
\caption{Profile log-likelihood for the three parameters in THO-B02B}
\label{Fig.7  }
\end{figure}

\begin{figure}[H]
\centering
\includegraphics[scale=0.5]{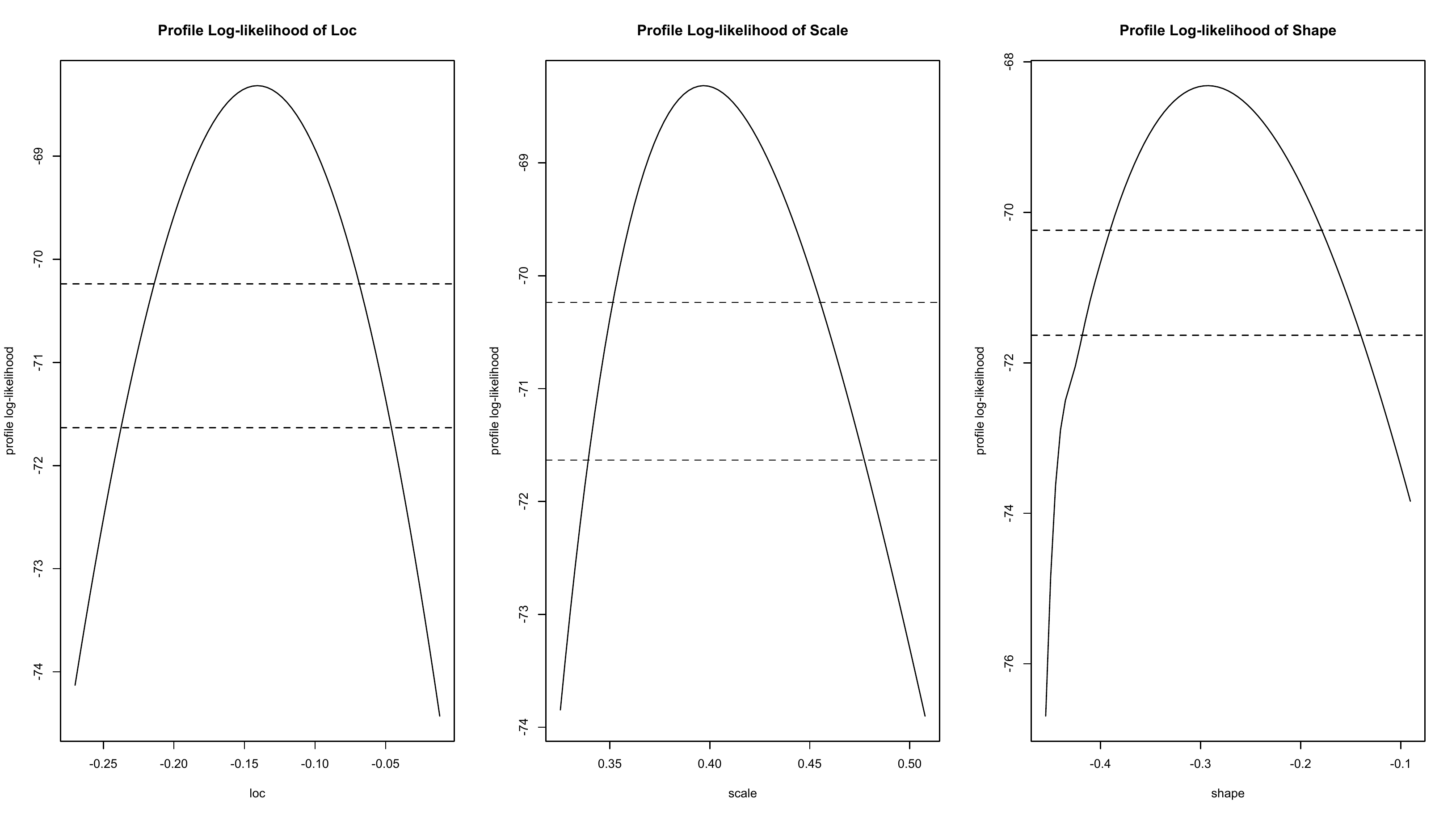}
\caption{Profile log-likelihood for the three parameters in THO-B02B}
\label{Fig.8  }
\end{figure}

\begin{figure}[H]
\centering
\includegraphics[scale=0.5]{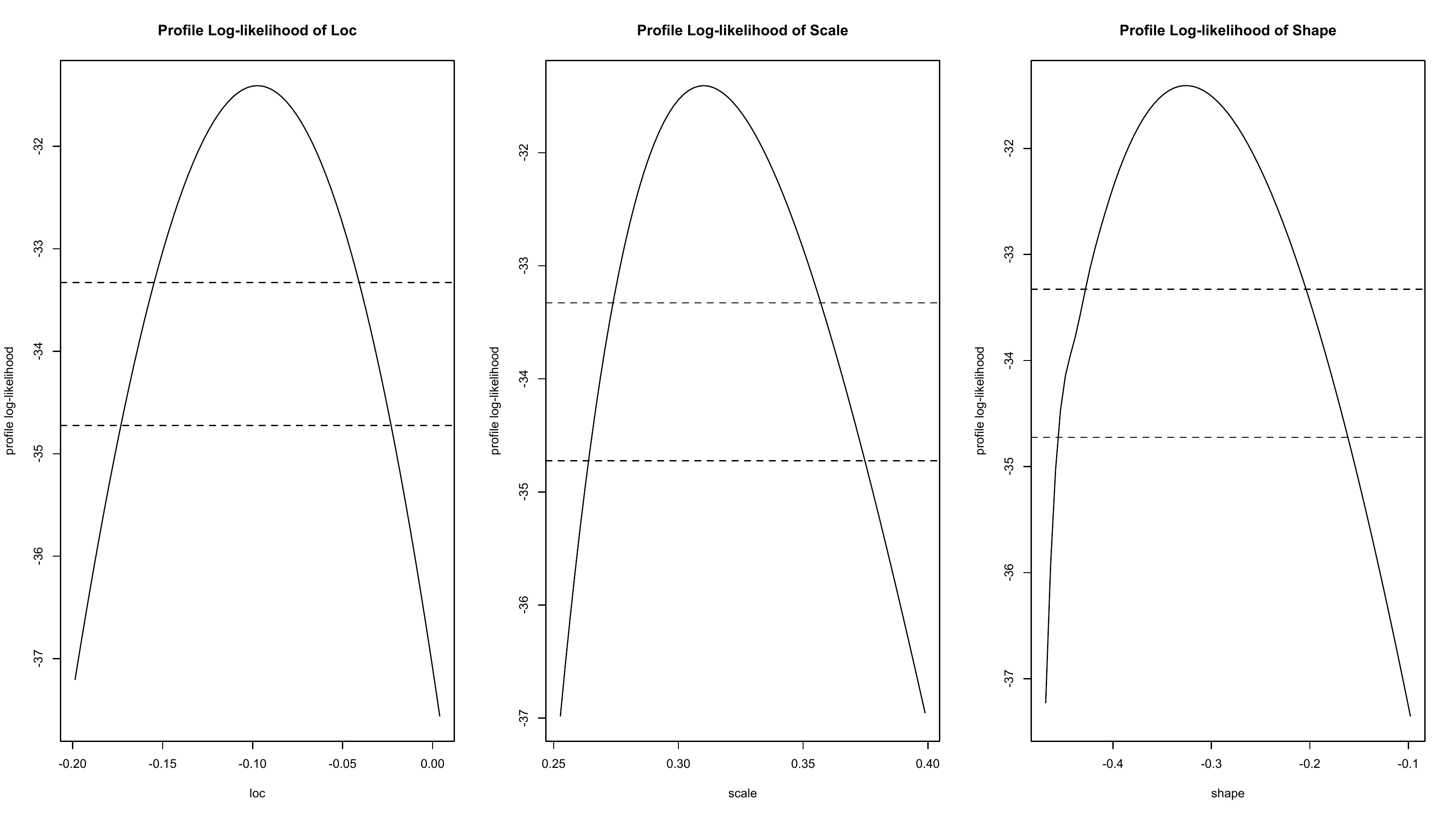}
\caption{Profile log-likelihood for the three parameters in THO-B03B}
\label{Fig.9  }
\end{figure}

\begin{figure}[H]
\centering
\includegraphics[scale=0.5]{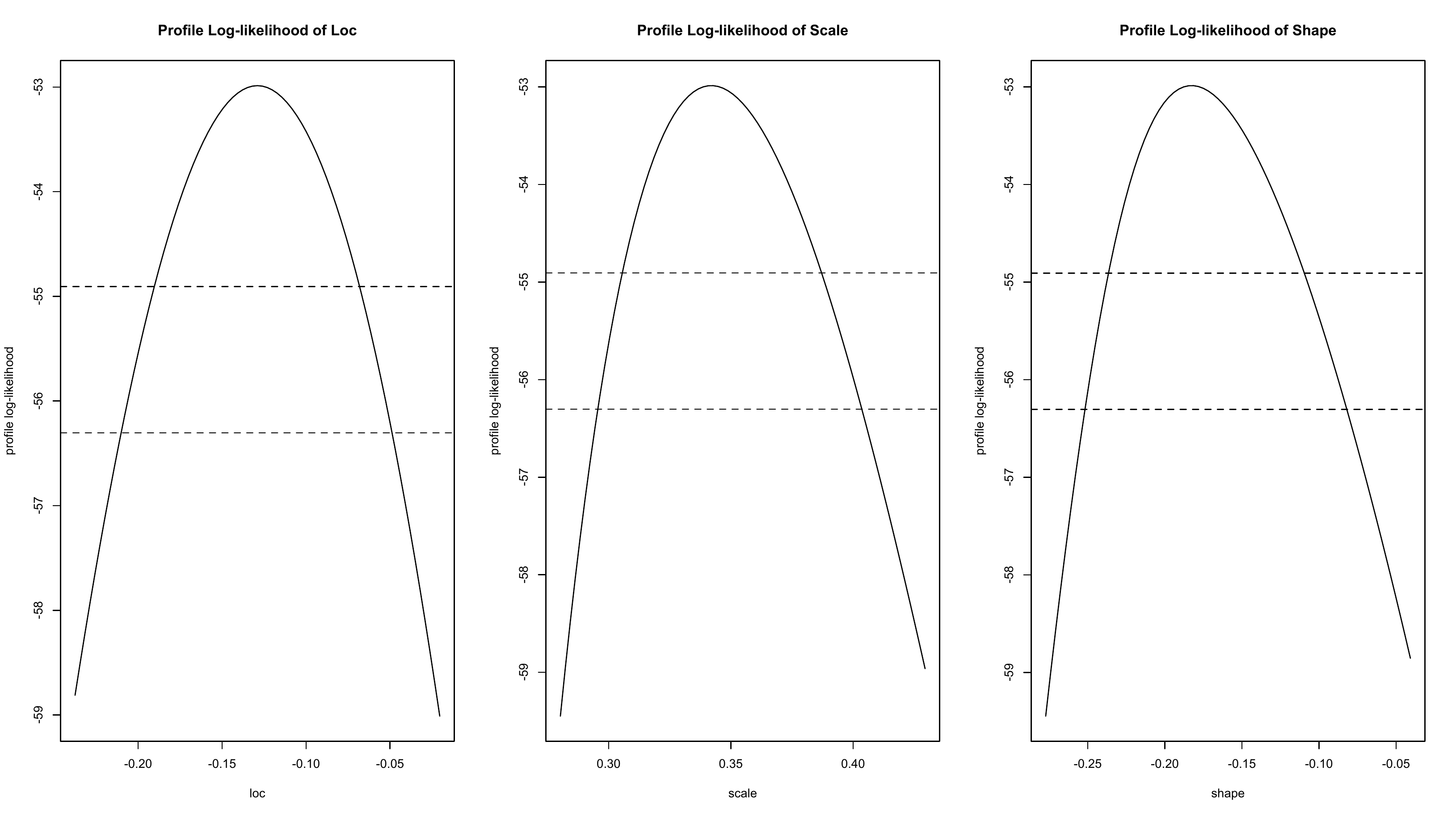}
\caption{Profile log-likelihood for the three parameters in THO-B04A}
\label{Fig.10 }
\end{figure}

\begin{figure}[H]
\centering
\includegraphics[scale=0.5]{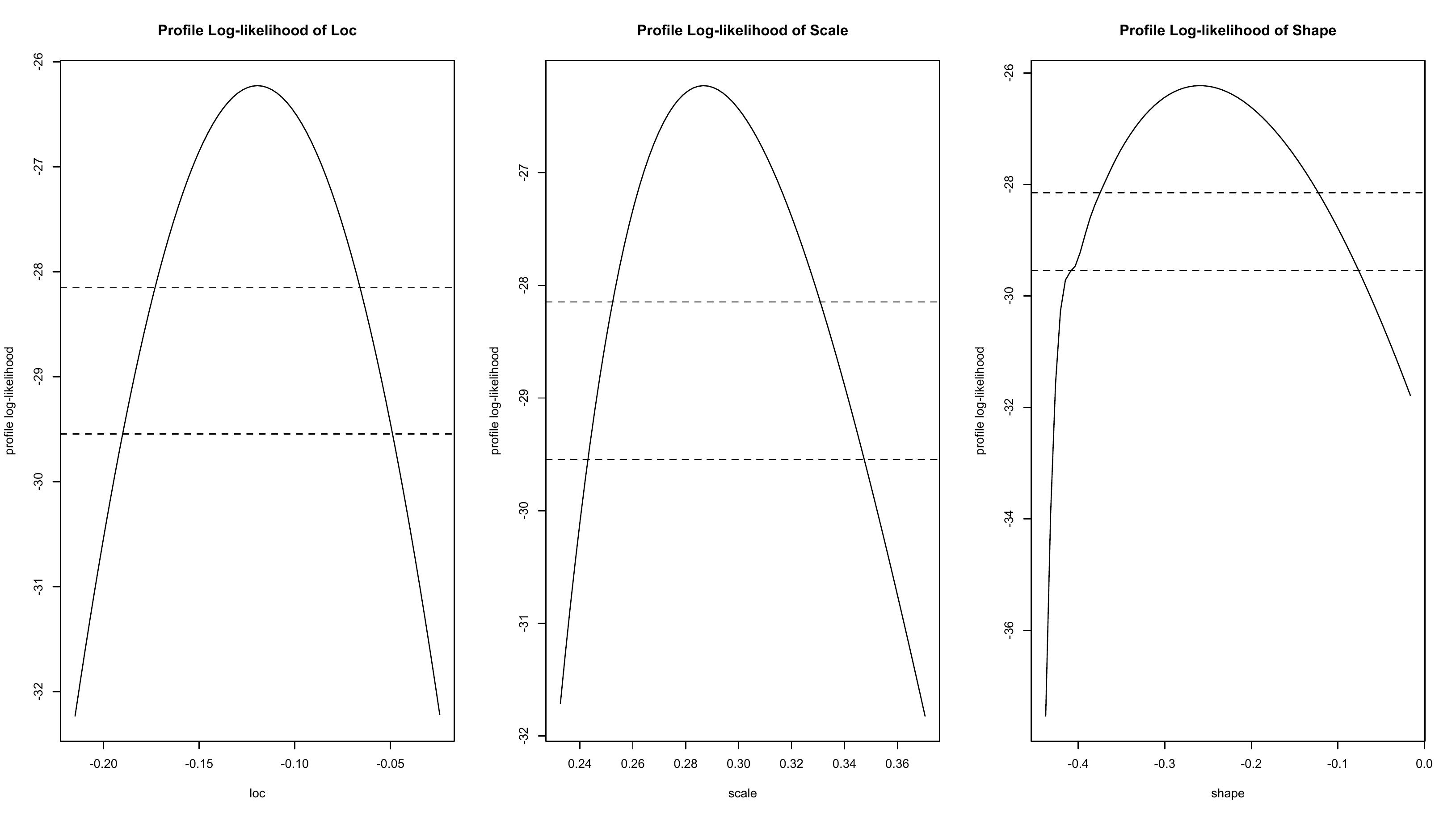}
\caption{Profile log-likelihood for the three parameters in THO-B05A}
\label{Fig.11 }
\end{figure}

\begin{figure}[H]
\centering
\includegraphics[scale=0.5]{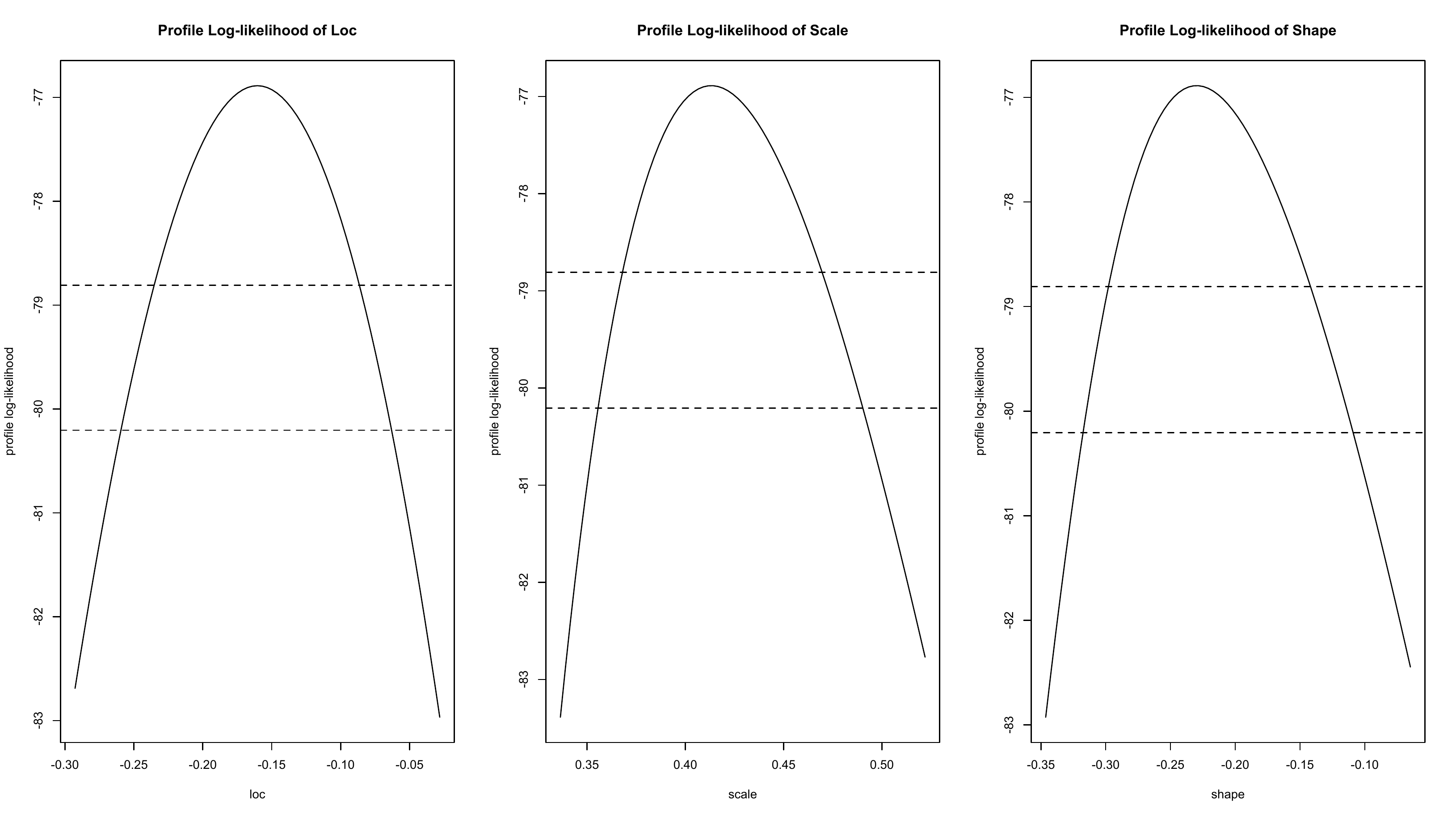}
\caption{Profile log-likelihood for the three parameters in THO-O01C}
\label{Fig.12 }
\end{figure}

\begin{figure}[H]
\centering
\includegraphics[scale=0.5]{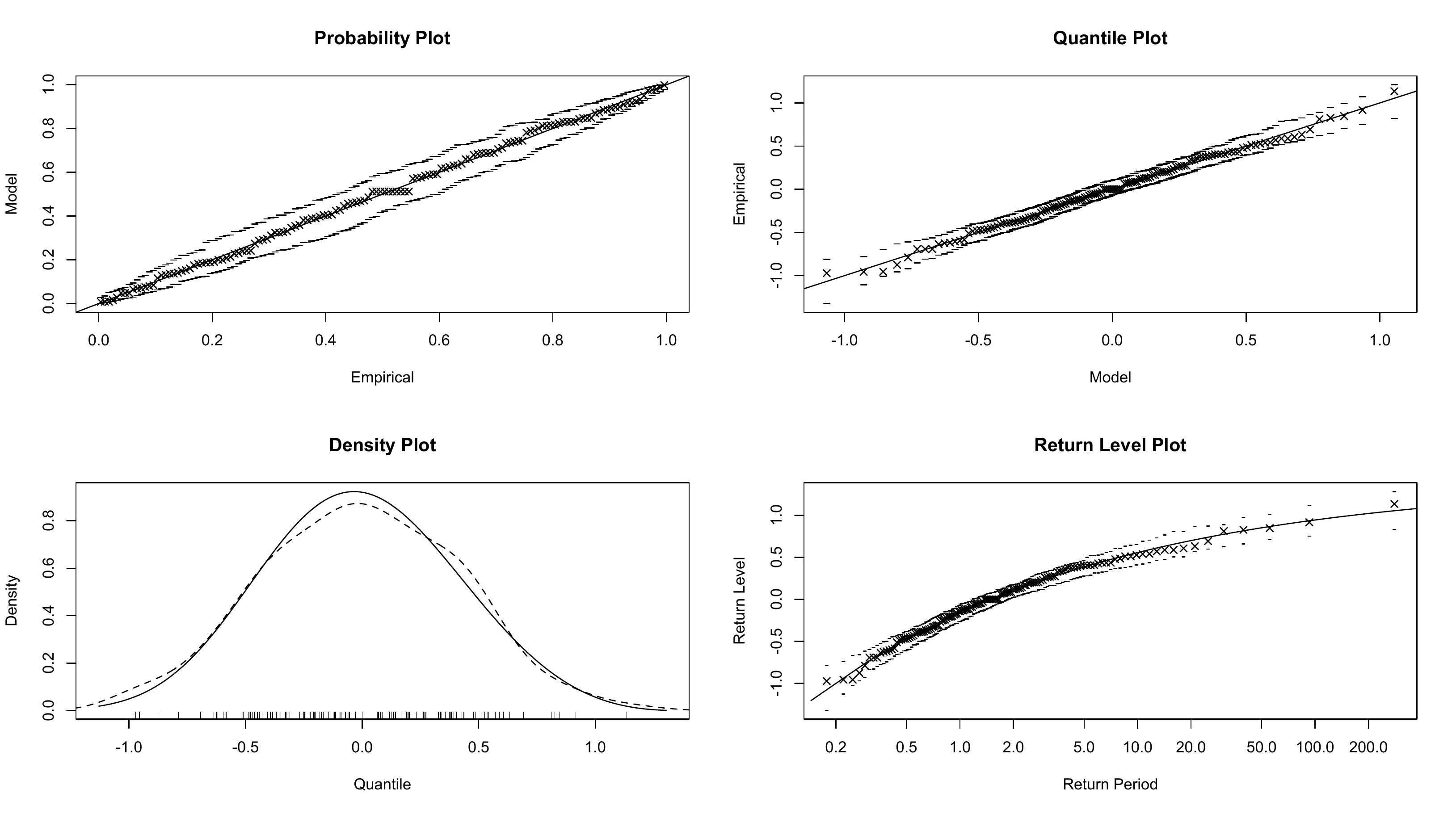}
\caption{Diagnostic plots for the EV fit to THO-A01B}
\label{Fig.13 }
\end{figure}

\begin{figure}[H]
\centering
\includegraphics[scale=0.5]{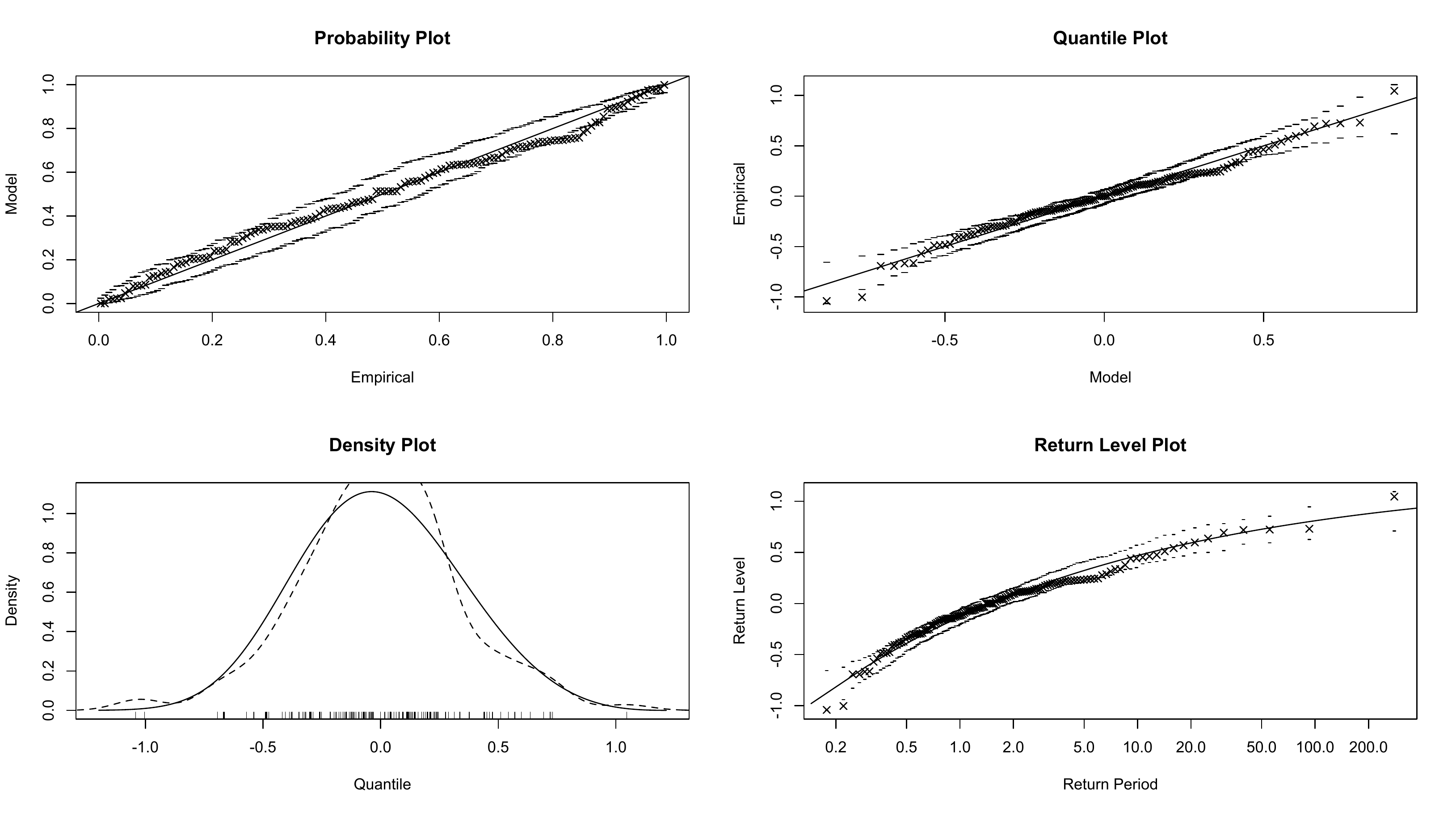}
\caption{Diagnostic plots for the EV fit to THO-A02A}
\label{Fig.14 }
\end{figure}

\begin{figure}[H]
\centering
\includegraphics[scale=0.5]{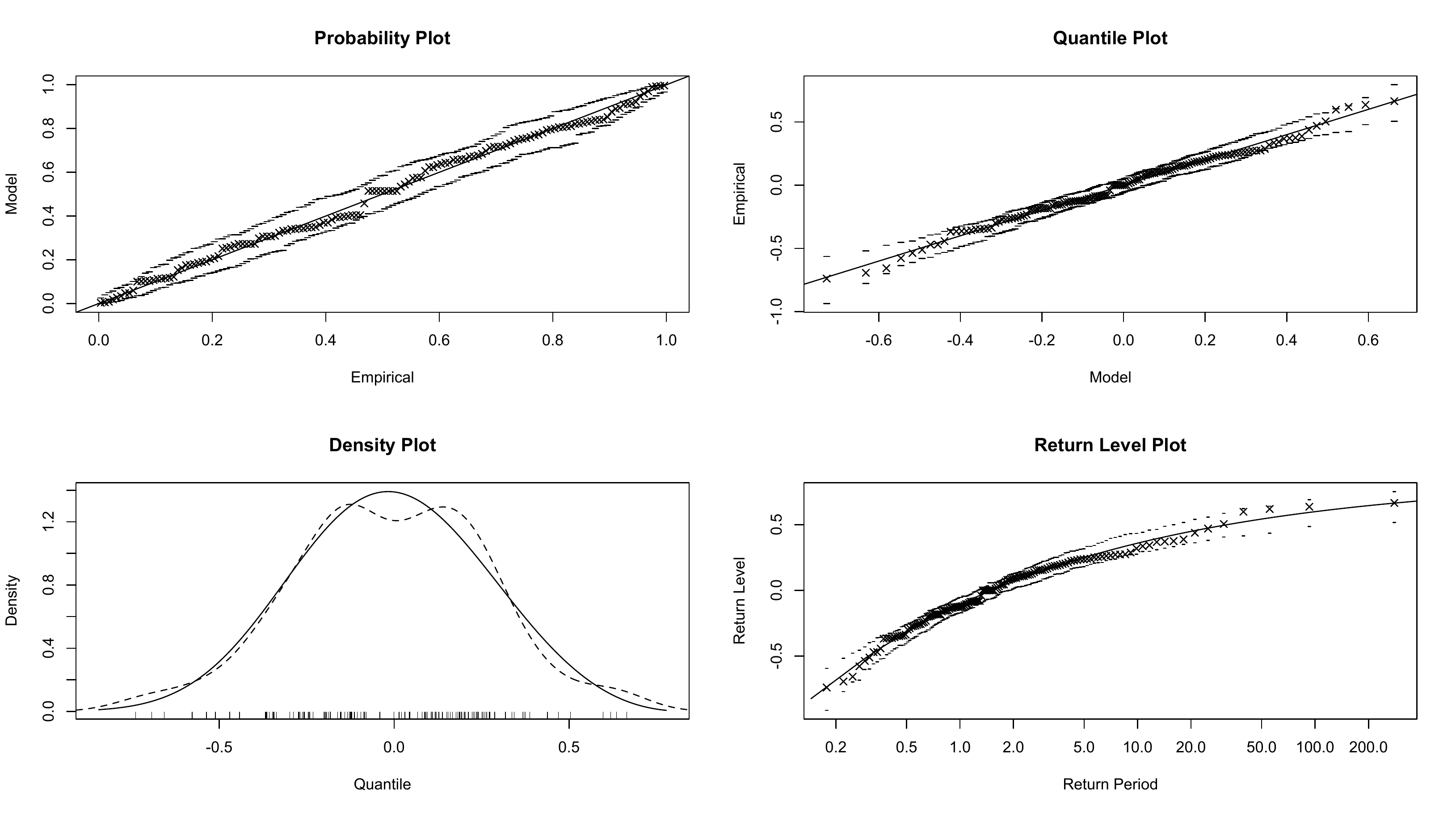}
\caption{Diagnostic plots for the EV fit to THO-A03A}
\label{Fig.15 }
\end{figure}

\begin{figure}[H]
\centering
\includegraphics[scale=0.5]{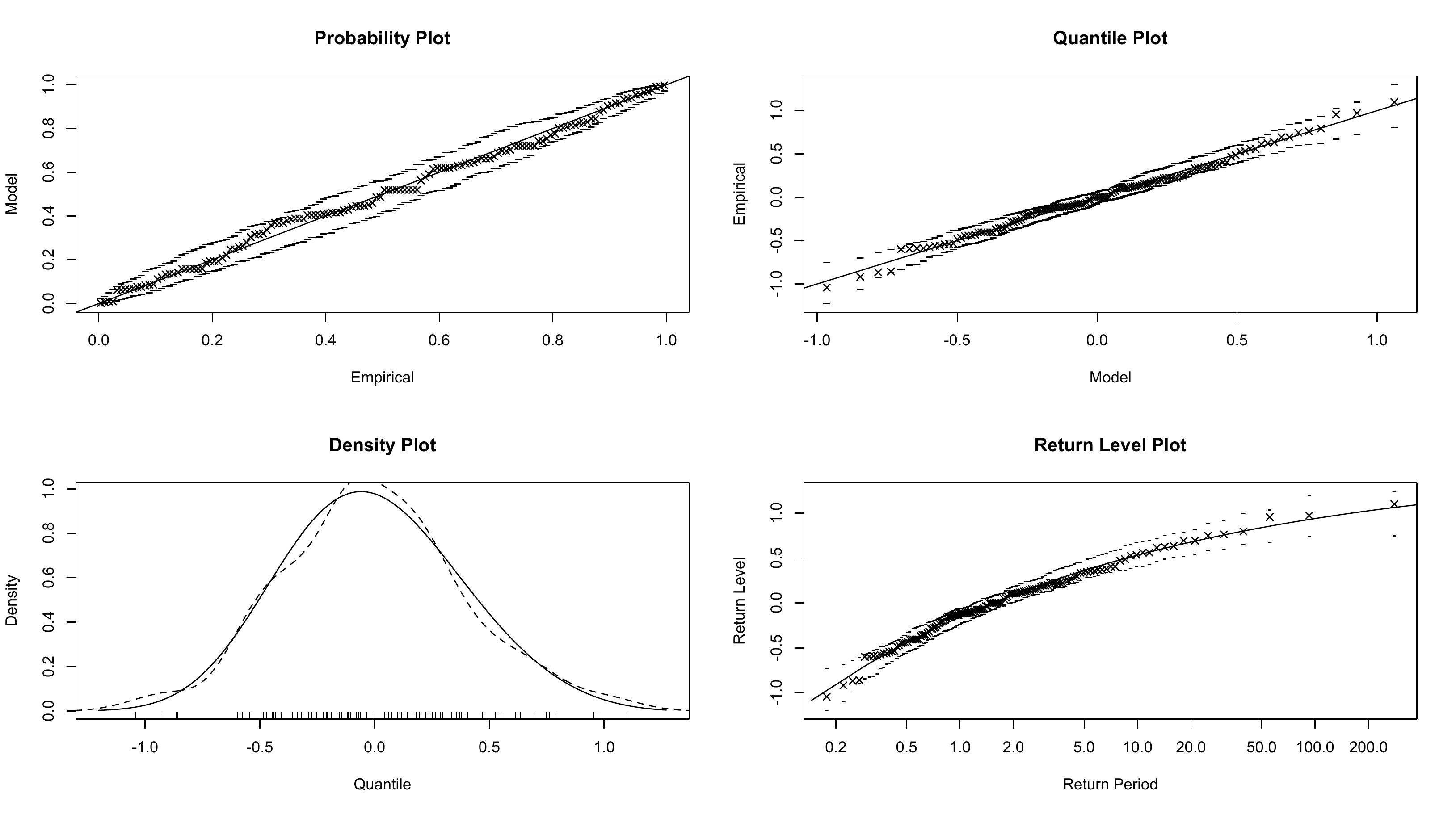}
\caption{Diagnostic plots for the EV fit to THO-B01A}
 \label{Fig.16}
\end{figure}

\begin{figure}[H]
\centering
\includegraphics[scale=0.5]{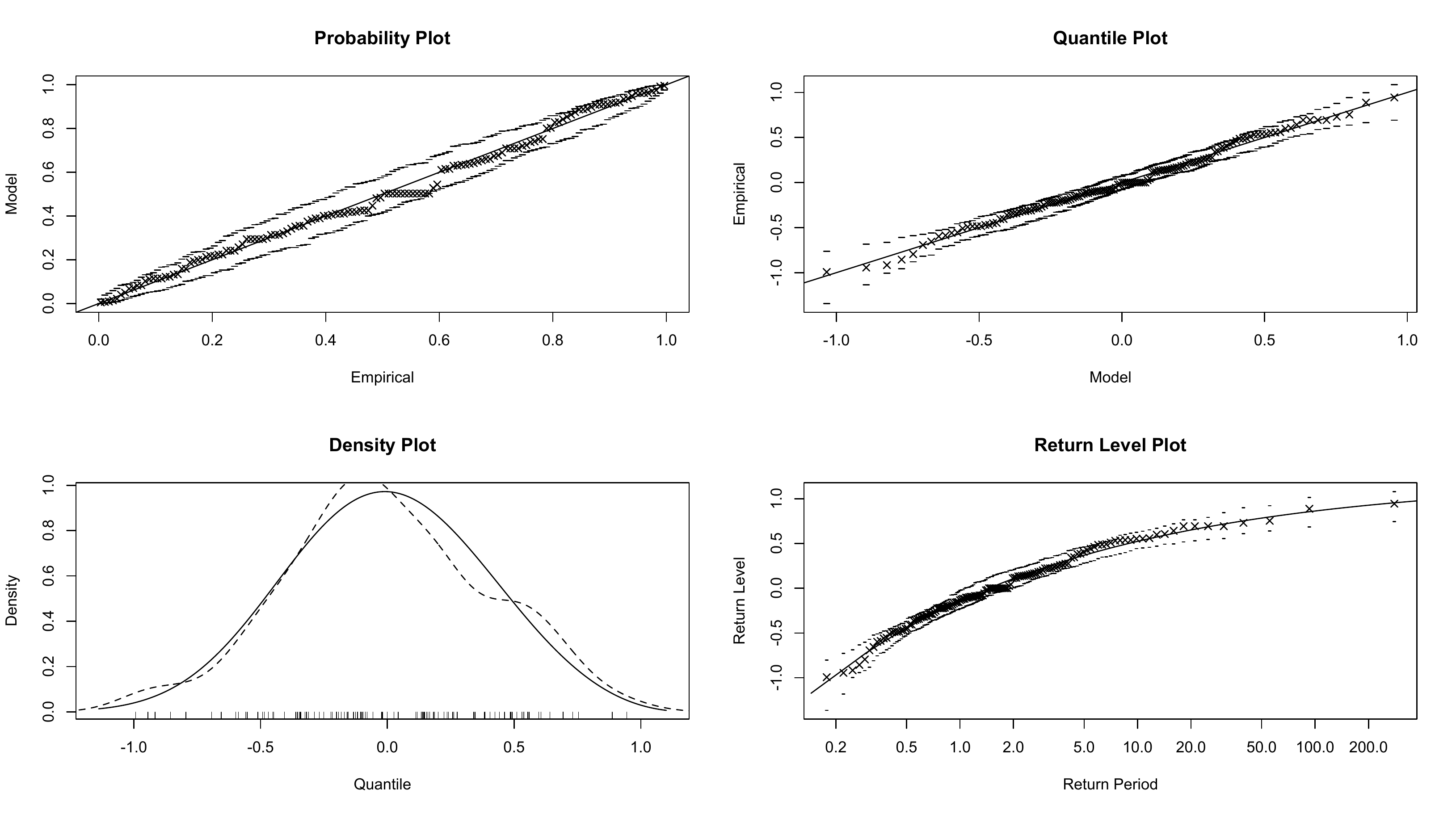}
\caption{Diagnostic plots for the EV fit to THO-B02B}
\label{Fig.17 }
\end{figure}

\begin{figure}[H]
\centering
\includegraphics[scale=0.5]{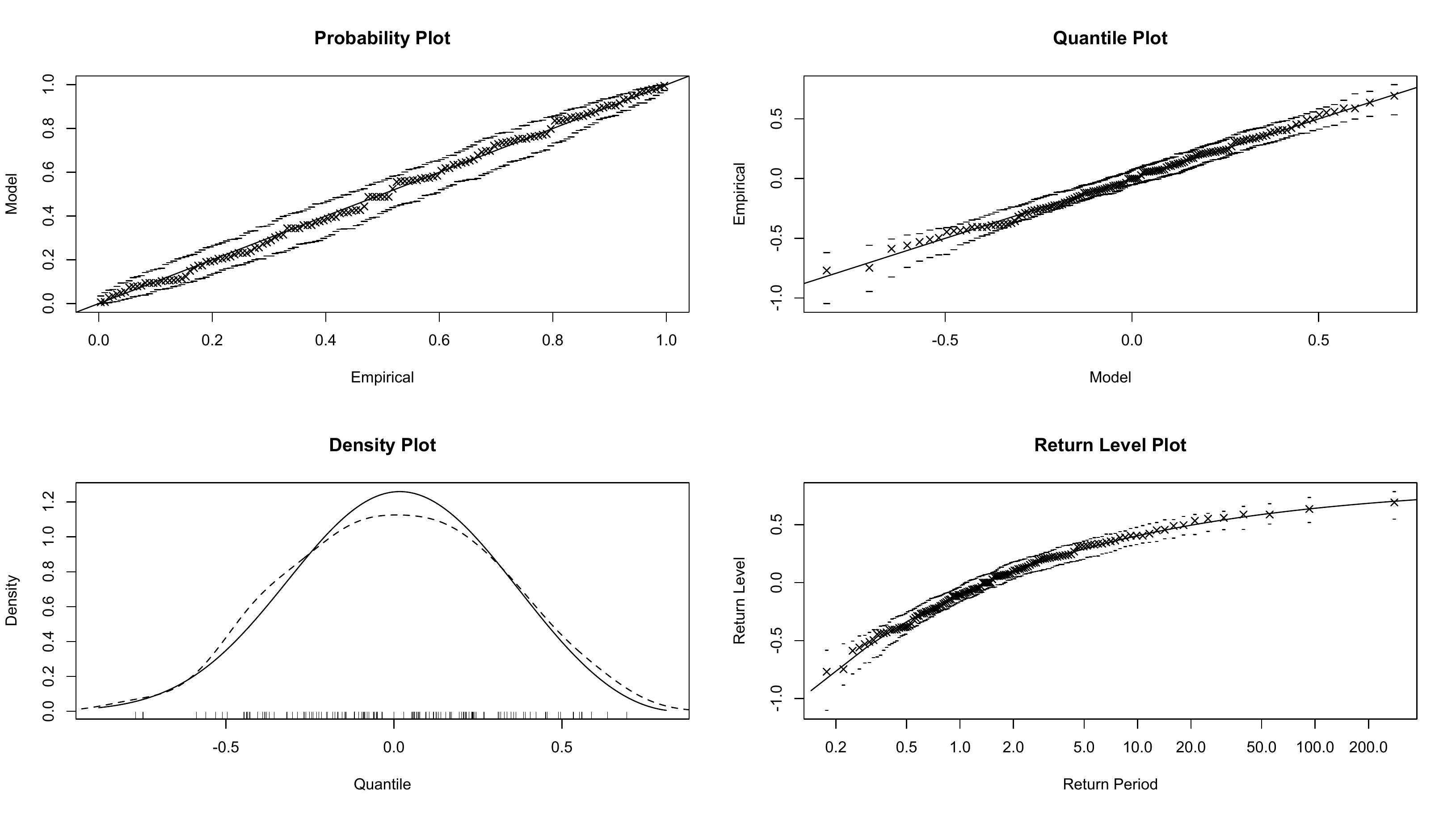}
\caption{Diagnostic plots for the EV fit to THO-B03B}
\label{Fig.18 }
\end{figure}

\begin{figure}[H]
\centering
\includegraphics[scale=0.5]{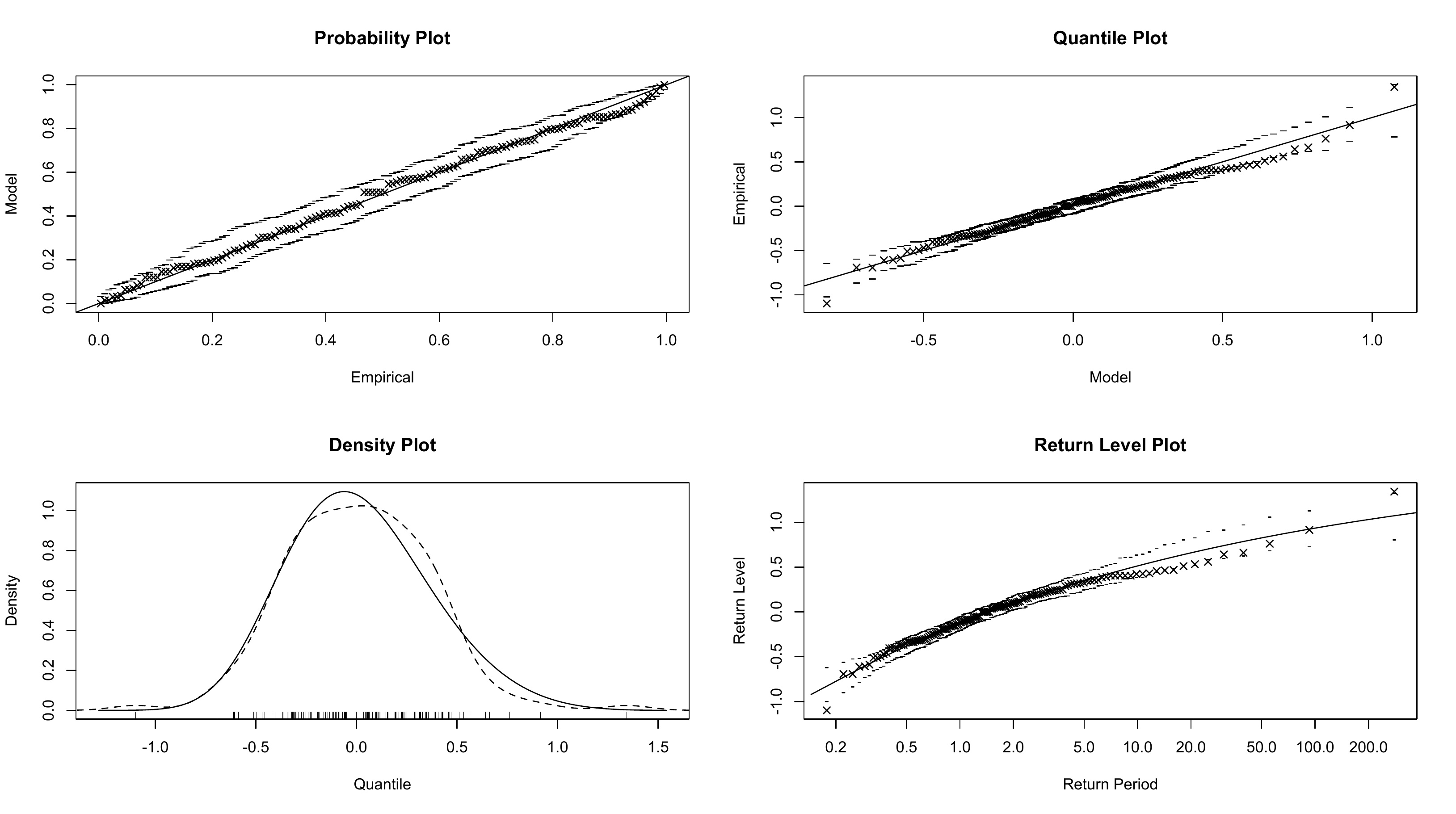}
\caption{Diagnostic plots for the EV fit to THO-B04A}
\label{Fig.19 }
\end{figure}

\begin{figure}[H]
\centering
\includegraphics[scale=0.5]{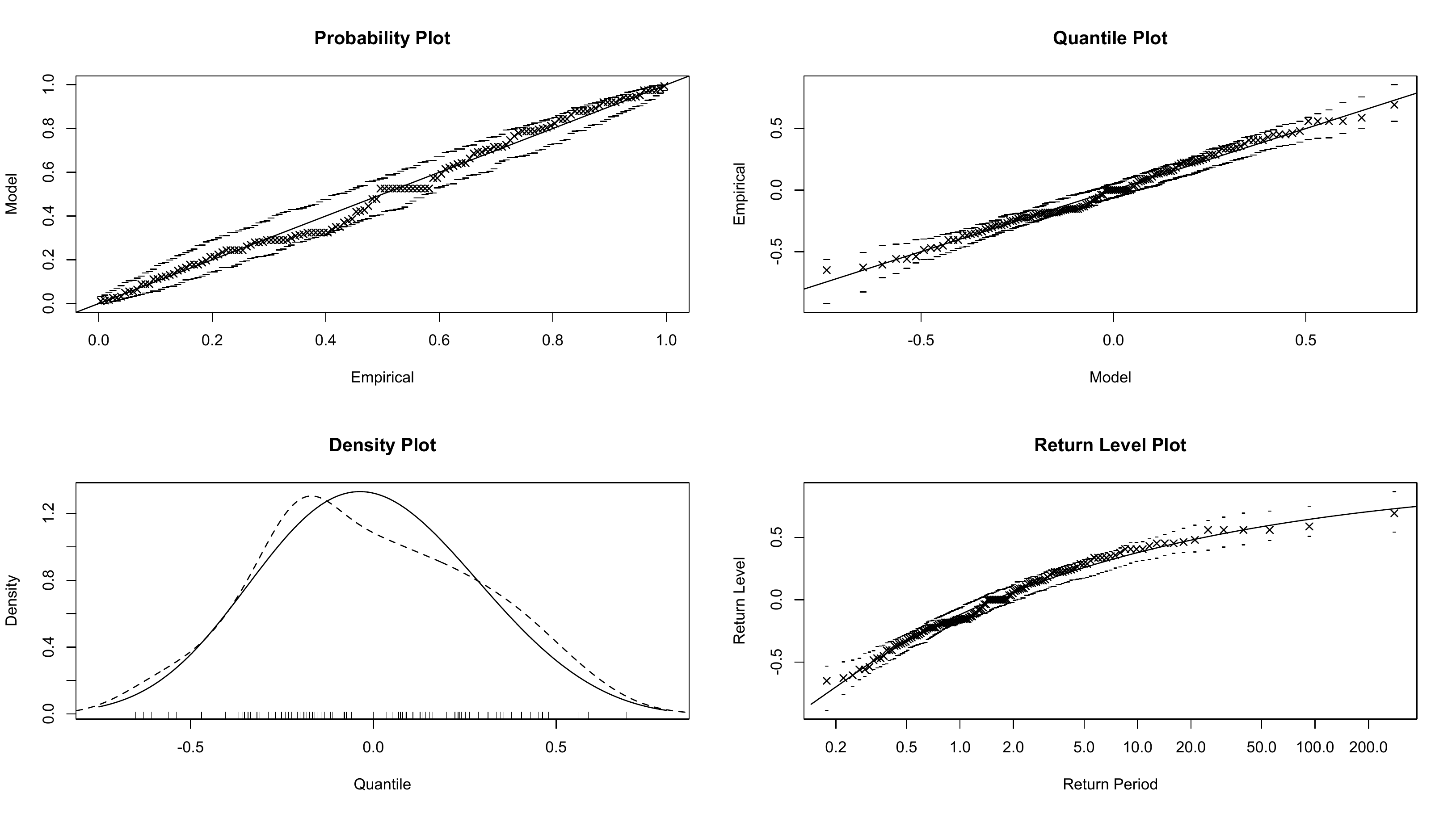}
\caption{Diagnostic plots for the EV fit to THO-B05A}
\label{Fig.20}
\end{figure}

\begin{figure}[H]
\centering
\includegraphics[scale=0.5]{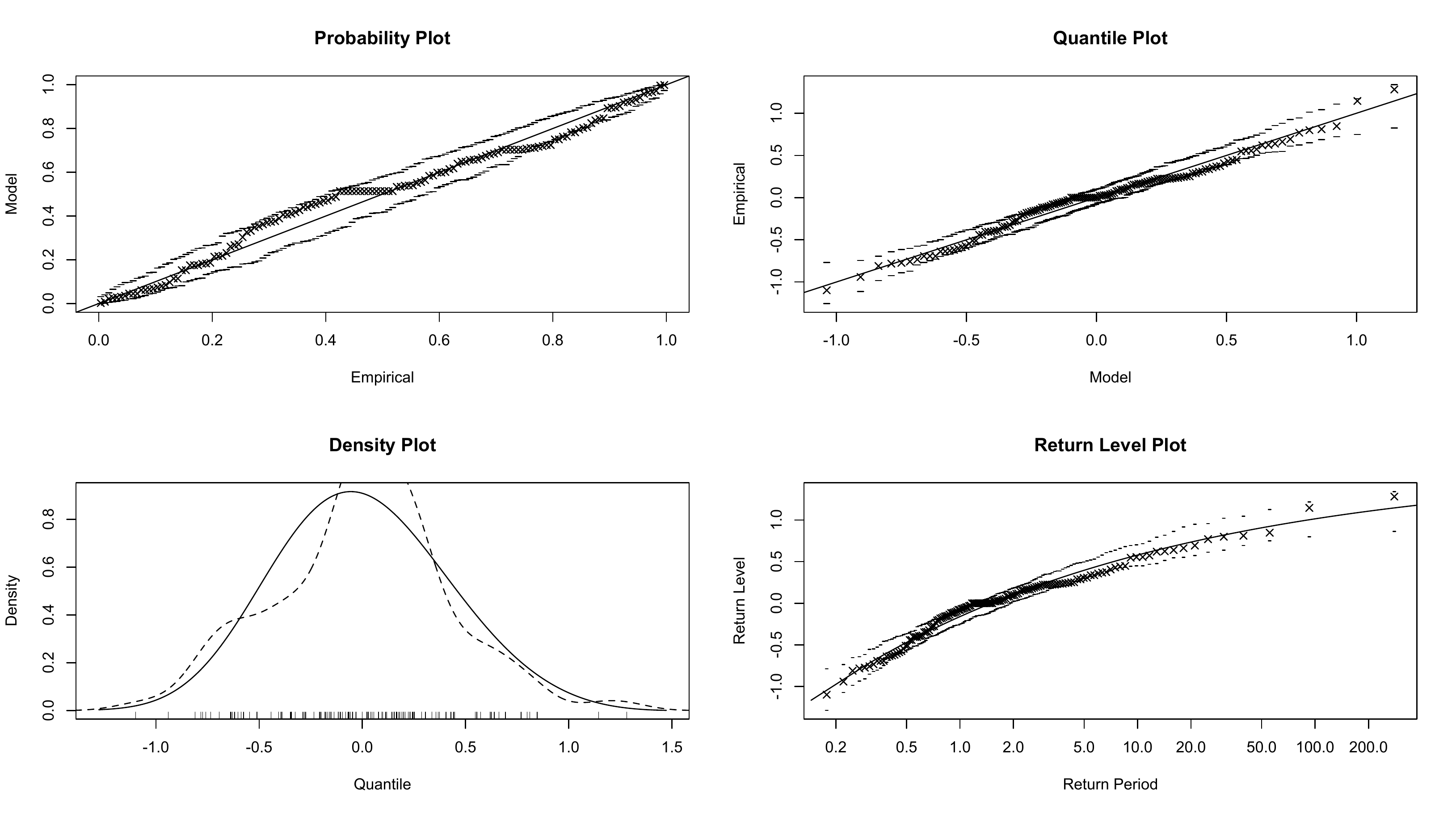}
\caption{Diagnostic plots for the EV fit to THO-O01C}
\label{Fig.21}
\end{figure}

\begin{itemize}
\item Gumble $\wedge(z)= \exp \Big (-\exp(-z)\Big)=EV_0(z) \quad z \in R, \quad \gamma=0$
\item Frechet $\Phi(z)= \exp\Big((-z)^{\dfrac{-1}{\gamma}} \Big) = EV_\gamma \Big (\dfrac{z-1}{\gamma} \Big ), \quad z > 0, \quad \gamma > 0$
\item Weibull $\Psi_\gamma (z)= \exp \Big (-(-z)^{\dfrac{1}{\gamma}} \Big )= EV_\gamma \Big (\dfrac{-z-1}{\gamma} \Big ), \quad z < 0, \quad \gamma <0$
\end{itemize}
One of us has shown in his investigations that the data are stationary \cite{Oazeley}. We fit an extreme value to the tree rings dataset. We  obtain the maximum likelihood fitting for the extreme value distribution. From Table \ref{table5}, we can see that all trees in the study have the shape of negative $\hat{\gamma}$. This would correspond to a bounded distribution. We carry out a Wald confidence interval for all the parameters. The Webull distribution is then seen as a possible candidate to model tree ring data. We examine our fitted models by using diagnostics plots to assess the accuracy of the extreme values. Both probability and quantile plots show reasonable extreme values fitting. The plots are almost linear; however, the return level plots are not linear, showing a slight convexity as can be seen in Figures \ref{Fig.4}, \ref{Fig.5 }, \ref{Fig.6 }, \ref{Fig.7  }, \ref{Fig.8  }, \ref{Fig.9  }, \ref{Fig.10 }, \ref{Fig.11 }, \ref{Fig.12 }, \ref{Fig.13 }, \ref{Fig.14 }, \ref{Fig.15 }, \ref{Fig.16}, \ref{Fig.17 }, \ref{Fig.18 }, \ref{Fig.19 }, \ref{Fig.20}, and \ref{Fig.21}.

\begin{table}[H]
\caption{Maximum likelihood fitting for the extreme value distribution }\centering
\begin{tabular*}{0.95\textwidth}{@{\extracolsep{\fill}}  c c  c  c  c  }
\hline
Sample &$\hat{\mu}$ (locution) &$\hat{\sigma}$ ( scale ) &$\hat{\gamma}$ (shape) & Standard Errors for $\hat{\gamma}$ \\
\hline
THO-A01B & -0.1576 &0.4145& -0.2654& 0.04540   \\
\hline
THO-A02A & -0.1319 &  0.3424   & -0.2471 & 0.03363 \\
\hline
THO-A03A & -0.1068  & 0.2770 & -0.2888  & 0.04648   \\
\hline
THO-B01A &  -0.1547  &  0.3831   &  -0.2275 & 0.04681   \\
\hline
THO-B02B & -0.1408 & 0.3969 & -0.2928 &0.05398 \\
\hline
THO-B03B &-0.09735 & 0.31012 &-0.32587 &  0.05699     \\
\hline
THO-B04A & -0.1290 &0.3420  &-0.1827   &  0.03155  \\
\hline
THO-B05A &  -0.1196& 0.2868 & -0.2593 & 0.06486  \\
\hline
THO-O01C & -0.1604 &  0.4132   & -0.2298  & 0.03884  \\
\hline
\end{tabular*}
\label{table5}
\end{table}

\clearpage 
%\begin{landspace}
 


 




%\end{landspace}

\clearpage
\section{Conclusion}
In this paper, we presented an innovative approach to analysing data using the machine learning algorithm to count tree rings, and showing the differences between the tree indexes' time series. The ARMA model, for time series, has been used to analyse the data in this work. A simulation study has been conducted and assessed to determine how the algorithm could use the true model correctly. The proportion to which the true model was selected by the Akika Information Criterion (AIC) was increased by increasing the ARMA model parameters. Moreover, the k-nearest neighbors (KNN) algorithm was used to determine the accuracy of the estimated models using a machine learning algorithm. It is noticeable that the random forest was the most accurate algorithm with lowest Root Mean Square Error (4.220). It can be concluded that the ML algorithm was able to yield a valuable and accurate analysis. The Weibull distribution is then a possible candidate to model tree rings data. We examine our fitted by using diagnostics plots for assessing the accuracy of the extreme model. Both probability and quantile plot show a reasonable extreme value fit, the plot are almost linear. The return level plots is not linear and shows a slight convexity. \\
In terms of future recommendations, the current study can be extended to evaluate the performance of our suggested approaches to different types of datasets, e.g., biomedical, chemometrics, and signal processing datasets. Non-parametric estimation can also be used to investigate the extreme value, in particular to derive lower bounds to the accuracy of non-parametric estimation of the extreme index.\\

\section*{Acknowledgements}

\section*{Declaration of Competing Interest}

The authors declare that they have no known competing financial interests or personal relationships that could have appeared to influence the work reported in this paper.

\end{document}